\documentclass[10pt,twocolumn,letterpaper]{article}

\usepackage{cvpr}              

\usepackage[dvipsnames]{xcolor}
\usepackage{listings}
\usepackage{amsmath}
\usepackage{booktabs}
\usepackage{multirow}
\usepackage{multicol}

\definecolor{cvprblue}{rgb}{0.21,0.49,0.74}
\usepackage[pagebackref,breaklinks,colorlinks,citecolor=cvprblue]{hyperref}

\def\paperID{*****} 
\def\confName{CVPR}
\def\confYear{2024}

\title{SPC-NeRF: Spatial Predictive Compression for Voxel Based Radiance Field}

\author{Zetian Song\\
Peking University\\
BeiJing, China\\
{\tt\small songzt@pku.edu.cn}
\and
Wenhong Duan\\
Shanghai Jiao Tong University\\
Shanghai, China\\
{\tt\small whduan@sjtu.edu.cn}
\and
Yuhuai Zhang\\
Peking University\\
BeiJing, China\\
{\tt\small zhangyuhuai@pku.edu.cn}
\and
Shiqi Wang\\
City University of Hong Kong\\
Hong Kong, China\\
{\tt\small shiqwang@cityu.edu.hk}
\and
Siwei Ma\\
Peking University\\
BeiJing, China\\
{\tt\small swma@pku.edu.cn}
\and
Wen Gao\\
Peking University\\
BeiJing, China\\
{\tt\small wgao@pku.edu.cn}
}

\begin{document}
\maketitle
\aboverulesep=0pt
\belowrulesep=0pt
\begin{abstract}
Representing the Neural Radiance Field (NeRF) with the explicit voxel grid (EVG) is a promising direction for improving NeRFs. However, the EVG representation is not efficient for storage and transmission because of the terrific memory cost. Current methods for compressing EVG mainly inherit the methods designed for neural network compression, such as pruning and quantization, which do not take full advantage of the spatial correlation of voxels. Inspired by prosperous digital image compression techniques, this paper proposes SPC-NeRF, a novel framework applying spatial predictive coding in EVG compression. The proposed framework can remove spatial redundancy efficiently for better compression performance.
Moreover, we model the bitrate and design a novel form of the loss function, where we can jointly optimize compression ratio and distortion to achieve higher coding efficiency. Extensive experiments demonstrate that our method can achieve 32\% bit saving compared to the state-of-the-art method VQRF on multiple representative test datasets, with comparable training time.
\end{abstract}    
\section{Introduction}
Generating models to represent three-dimensional (3D) scenes and distributing them as photos on the Internet can profoundly revolutionize entertainment and future media. 
The emergence of Neural Radiance Field (NeRF)~\cite{10.1007/978-3-030-58452-8_24} has demonstrated promising results in reconstructing 3D scenes from sparse views and rendering 
photo-realistic pictures from arbitrary viewpoints.

Primary NeRF models in existing literature tend to employ neural networks for implicit 3D scene modeling. However, these methods suffer from high computational complexity, resulting in heavy training and rendering burdens. Recent research has established a promising direction to accelerate the computation of training and rendering by learning voxel-based representations for the radiance field~\cite{sun2022direct, fridovich2022plenoxels}, known as Explicit Voxel Grid (EVG) representations. The EVG representations 
express significant potential to mitigate the computational complexity challenges. Nevertheless, the EVG representations will generate extra huge parameters, significantly larger than neural network based representations. The enormous parameters of the voxel grid lay heavy storage and computational burdens, hindering further development and actual application.

\begin{figure}[t]
    \centering
  \begin{subfigure}{0.48\linewidth}
    \includegraphics[width=\linewidth]{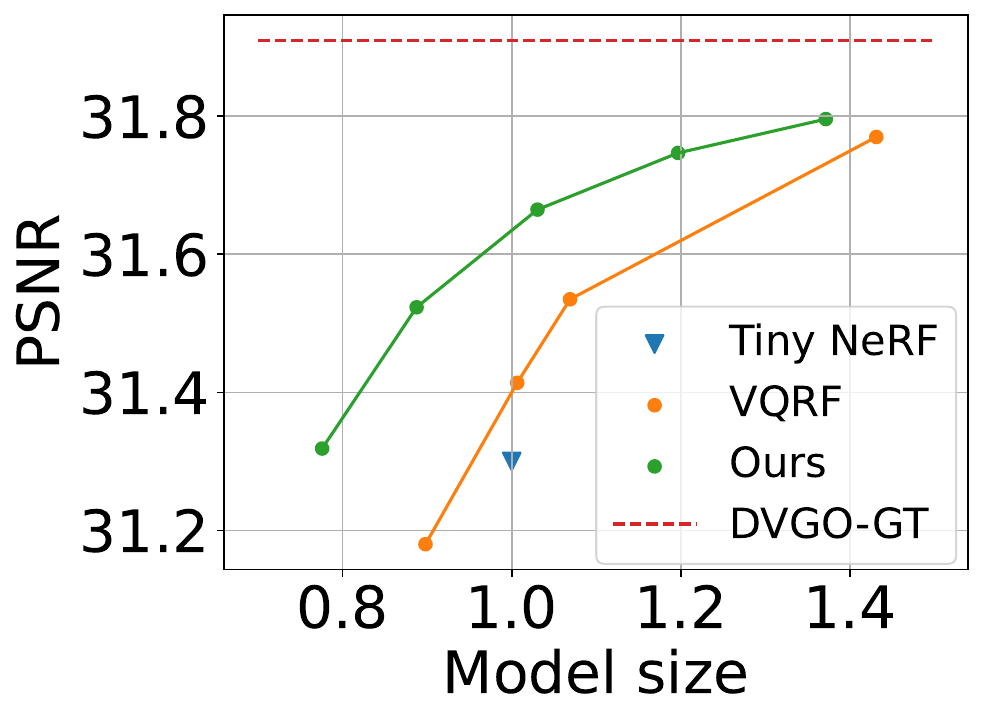} 
    \caption{Voxel resolution $160^3$.}
  \end{subfigure}
  \hfill
  \begin{subfigure}{0.49\linewidth}
    \includegraphics[width=\linewidth]{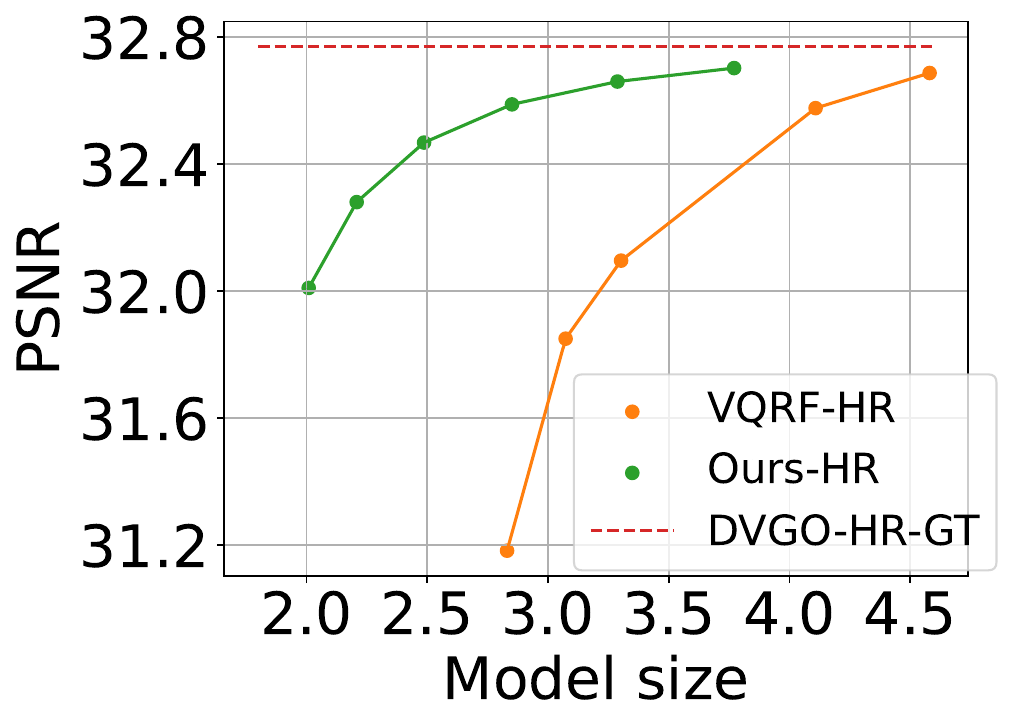} 
    \caption{Voxel resolution $256^3$.}
  \end{subfigure}
\caption{
Rate-Distortion curves of our method compared to state-of-the-art EVG NeRF compression methods on the Synthetic NeRF dataset.
}
\label{overall perfomance}
\end{figure}
To tackle these challenges, recent works~\cite{zhao2023tinynerf, deng2023compressing, li2023compressing} applied data compression methods for EVG-based NeRFs, including importance pruning, frequency domain transformation and vector quantization. These approaches preserve the performance of the original model as much as possible while reducing the model size. It is observed that around 100$\times$ compression can be realized with negligible loss of rendering quality. In addition, Wang \emph{et al.} proposed a motion compensation method for better compressing dynamic NeRF~\cite{wang2023neural}.
Although effective, these methods rarely take into account the spatial correlation of the EVG data. Furthermore, these works train the voxel grid with little regard for compression due to the complexity of modeling actual coding bits in such cases.

In this paper, we perform a detailed analysis of the spatial correlation between adjacent voxels, which shows that substantial spatial redundancy exists in the voxel grid. To address the aforementioned issue, we propose SPC-NeRF, a Rate-Distortion~(RD) optimized predictive coding framework for compressing the EVG represented NeRF.  The proposed method can take full advantage of the spatial correlation and remove the spatial redundancy to achieve a higher compression ratio. Moreover, our method can jointly optimize the rendering distortion and the coding bitrate based on the RD loss function, which can make a trade-off between quality and model size while training the voxel grid.
%
%

The major contributions of this work are summarized as follows,
\begin{itemize}
    \item We propose a generic pipeline for spatial predictive compression of the EVG NeRF, which can efficiently remove the spatial redundancy to obtain better compression performance.
    \item We introduce a novel weighted RD loss with a bit estimation method adapted to our framework. In addition, a two-step finetuning procedure is proposed to realize adaptive quantization precision.
    \item Our experimental results demonstrate that SPC-NeRF can achieve 32\% bit saving with a minor increase in training overhead compared to the state-of-the-art method.
\end{itemize}

\section{Related Work}
\subsection{Neural Radiance Field}
Neural radiance field~\cite{10.1007/978-3-030-58452-8_24} has shown great ability in 3D reconstruction, and motivated massive follow-up works, such as editing~\cite{yuan2022nerf, bao2023sine} and speeding-up~\cite{muller2022instant, cao2023hexplane}. Representing dynamic scenes, or Free Viewpoint Videos~(FVVs), with NeRF is also widely researched~\cite{pumarola2021d, fang2022fast, li2022neural}.

In addition to deep neural network based NeRFs, multiple representation methods for modeling the radiance field have been proposed. Sun \emph{et al.} proposed DVGO~\cite{sun2022direct}, an explicit voxel grid representation for the radiance field, with a shallow network to learn the color features. Fridovich \emph{et al.} proposed Plenoxels~\cite{fridovich2022plenoxels}, a method that goes beyond neural-based features by incorporating handcrafted features based on the spherical harmonic function.
Chen \emph{et al.} and Fridovich \emph{et al.} decomposed the 3D voxel grid and proposed a multi-plane representation for both static and dynamic NeRFs~\cite{chen2022tensorf, fridovich2023k}. These methods made a significant improvement in training and rendering speed compared to fully neural representations. However, their models tend to be large in size, making them unsuitable for extensive storage and distribution on the network. 

Recently, several methods have been proposed for compressing neural-based NeRF~\cite{bird20213d, 10095668}, EVG NeRF~\cite{deng2023compressing, li2023compressing, zhao2023tinynerf}, multi-plane NeRF~\cite{rho2023masked, tang2022compressible} and dynamic NeRF~\cite{wang2023neural}. Tang \emph{et al.} proposed CC-NeRF~\cite{tang2022compressible}, reducing model size via rank-residual decomposition. Deng \emph{et al.} proposed Re:NeRF~\cite{deng2023compressing}, a progressive pruning approach for EVG-represented NeRFs. Zhao \emph{et al.} and Rho \emph{et al.} applied orthogonal transformations to compress explicit NeRFs in the frequency domain~\cite{zhao2023tinynerf, rho2023masked}. 
Li et al. introduced VQRF~\cite{li2023compressing}, achieving SOTA compression performance for EVG NeRF. The VQRF initially performs importance pruning, removing unimportant voxels, and then conducts vector quantization on the majority of the remaining voxels. Finally, it jointly finetunes the remaining parameters and the codebook to compensate for quantization loss.
\\
\subsection{Predictive Coding}
Predictive coding~\cite{wiegand2011source} is a fundamental coding concept that involves calculating a prediction value for the current sample based on previous samples. Predictive coding encodes the residual (prediction error) instead of encoding the original sample value, aiming to eliminate redundancies. This approach was widely used in image and video coding standards such as JPEG-LS~\cite{855427}, H.266/VVC~\cite{bross2021overview}, AVS3~\cite{zhang2019recent} and AV1~\cite{han2021technical}. In order to deal with different types of redundancy, a variety of effective prediction methods have been developed. Among them, cross-component predictive coding~\cite{9190915} was designed to eliminate redundancy between color components. Temporal predictive coding utilized the similarity of adjacent temporal data, modeling the motion between timestamps to generate prediction value~\cite{bross2014inter}. The spatial predictive coding~\cite{pfaff2021intra} most relevant to this work predicted the current sample using spatially adjacent samples to explore spatial data correlations. In this work, we select one from seven previously encountered neighboring voxels as the prediction value for the current voxel, and we record the index of the reference for each voxel.

\subsection{Network compression}
Network compression has been extensively studied as an effective approach to reduce the computational and storage complexity of neural networks. Existing methods for network compression can be broadly categorized into four main types: low-rank factorization~\cite{swaminathan2020sparse, 8478366}, knowledge distillation~\cite{gou2021knowledge}, weight pruning~\cite{he2017channel, ma2021non, li2016pruning} and network quantization~\cite{choi2018pact, duan2022differential}. Since NeRF is a specialized neural network, many techniques proven effective for general neural networks have also demonstrated efficacy for NeRF, even for the EVG NeRF. TensorRF~\cite{chen2022tensorf} and CC-NeRF\cite{tang2022compressible} applied the low-rank decomposition approach to reduce the storage overhead of EVG NeRF. Several works adopted Weight pruning techniques~\cite{deng2023compressing, zhao2023tinynerf, xie2023hollownerf}. Besides, the model quantization was also employed to compress NeRFs~\cite{Gordon_2023_WACV, li2023compressing}.

\section{Problem Statement}
\begin{figure}[t]
    \centering
  \begin{subfigure}{0.47\linewidth}
    \includegraphics[width=\linewidth]{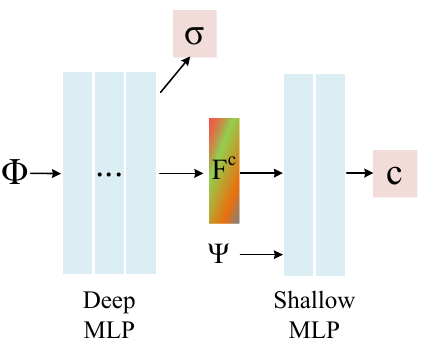} 
    \caption{NeRF}
  \end{subfigure}
  \hfill
  \begin{subfigure}{0.51\linewidth}
    \includegraphics[width=\linewidth]{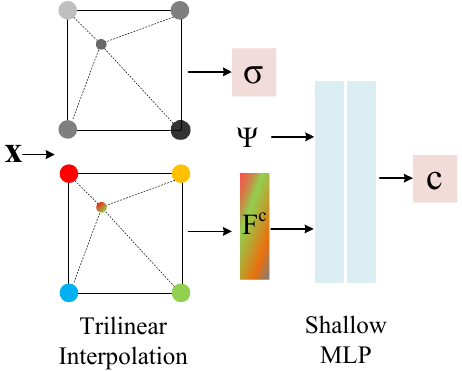} 
    \caption{EVG NeRF}
  \end{subfigure}
\caption{
Two different approaches to mapping 3D coordinates to color value $c$ and volume density $\sigma$. $\bf{x}$ denotes the input Cartesian coordinate, $\Phi$ and $\Psi$ denotes the 3D coordinate and the view direction of positional encoding, respectively. $F^c$ represents the color feature.
}
\label{nerf_abc}
\end{figure}

\begin{figure*}[t]
\centering
\includegraphics[width=0.98\textwidth]{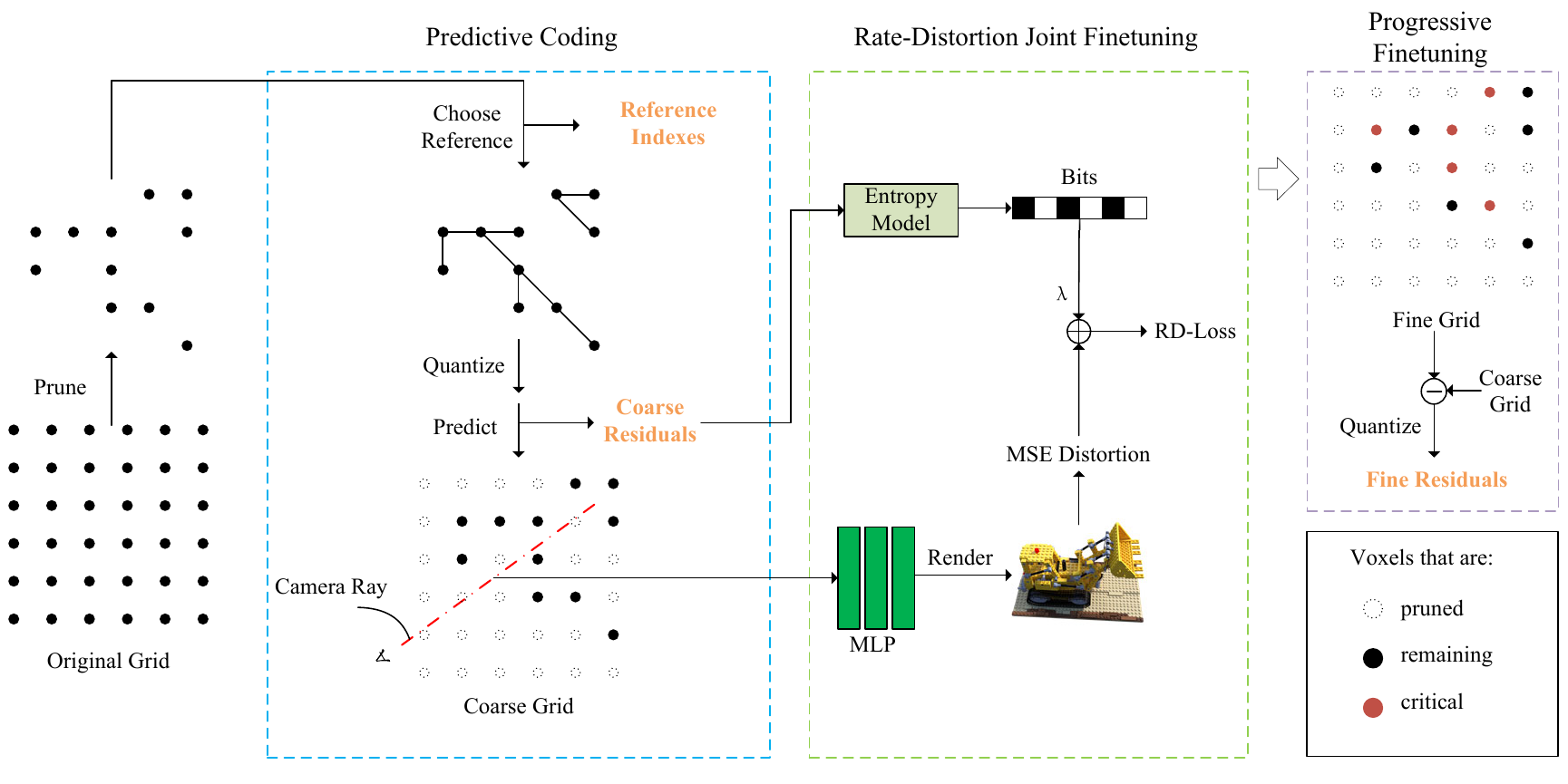} 
\caption{The overview framework of our method.
The orange font indicates the variables necessary to reconstruct the feature grid, i.e., the syntax elements that are written into the coding bitstream.}
\label{framework}
\end{figure*}
Neural radiance fields learn a function to map the 3D point $\bf{x}\in \mathbf{R}^3$ and the view direction $\bf{\tau}\in \mathbf{R}^2$ to the color value $c$ and the volume density $\sigma$. The pixel value can then be obtained according to the volume rendering method~\cite{max1995optical}. The original NeRF relies entirely on a Multi-Layer Perceptron~(MLP) to learn the mapping mentioned above. Although this approach has a relatively small model size, it comes with a high computational cost, making both training and rendering slow. Recent advancements employed the voxel grid to represent NeRF explicitly, constructing both a density grid and a feature grid~\cite{sun2022direct}. Instead of relying on the complicated forward inference of a deep MLP, this approach directly obtains the color feature and volumetric density through simple trilinear interpolation. The color feature is concatenated with the view direction and then fed into a shallow for RGB values. We depict the two different approaches in Fig. \ref{nerf_abc}

As a 3D scene reconstruction method, the performance of NeRF is usually measured by the distortion metrics between the rendered image and the ground truth image. At the same time, NeRF can also be regarded as a novel representation of a 3D free viewpoint visual system. In this sense, NeRF is similar to point clouds, meshes and light field data for 3D scenes, or images and videos for 2D visual signals.

Based on the above analysis, we view the NeRF compression as a problem of finding the most compact representation for a 3D visual system. Therefore, we define the NeRF compression problem as finding an efficient and compact representation that minimizes the bitrate $\mathcal{R}(\mathcal{D})$ for a given distortion, or minimizes the distortion for a given bitrate. The bitrate represents the size of NeRF model. The distortion can be defined as follows,
\begin{equation}
    \mathcal{D} = \mathbb{E}_{(\bf{x}, \bf{\tau})} d(\mathcal{I}_{ori}, \mathcal{I}_{rec}),
\end{equation}
where $d(\mathcal{I}_{ori}, \mathcal{I}_{rec})$ is the distortion metric between ground truth and the rendered image.
Since this definition is not computable, we follow the general configuration of NeRFs~\cite{10.1007/978-3-030-58452-8_24, sun2022direct}
to approximate the above distortion using the average distortion over sparse viewpoints.

\section{Rate-Distortion Optimized Spatial Predictive Coding Framework}
Before going into details, we first depict an overview of the proposed framework in Fig.~\ref{framework}. The proposed method is composed of a predictive coding module, an RD loss function and a two-step finetuning procedure. Specifically, the predictive coding module aims to remove spatial redundancy, the novel loss function we designed can make a trade-off between model size and rendering quality, and the two-step finetuning procedure minimizes the rate-distortion loss of the NeRF model.

We first apply the DVGO~\cite{sun2022direct} pipeline to obtain the uncompressed voxel grid. We prune about 90\% unimportant voxels and mark a small part of the remaining voxels critical. Then we select a reference for each unpruned voxel and construct a graph that records the reference relation. We apply a unique bit estimation method to form the RD loss and finetune the voxel grid, after which the actual quantization and prediction are performed to generate coarse residuals. Finally, we post finetune the model with all non-critical voxels frozen, generating fine residuals for reconstructing the critical voxels.

All syntax elements we code into the bitstream include the following parts: 1) reference indexes, 2) coarse residuals, 3) fine residuals, 4) prune mask, 5) critical mask, 6) non-prune density grid, 7) weights of the shallow MLP, 8) metadata. The residuals and indexes are entropy coded via arithmetic coding~\cite{witten1987arithmetic}, whilst other syntax elements that occupy little storage are coded with a simple LZ77~\cite{ziv1977universal} codec.

\subsection{Spatial Predictive Coding}
\begin{figure}[t]
\centering
\includegraphics[width=0.6\columnwidth]{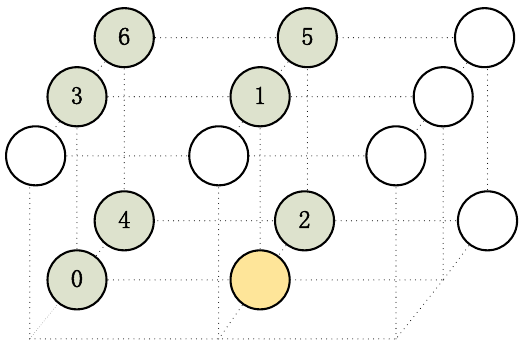} 
\caption{The yellow voxel denotes the current processing voxel; others are the available neighbors, wherein the green voxels represent the reference candidates we used with their indexes, i.e., voxels from which we select reference.}
\label{refs}
\end{figure}
Despite the different data types between the feature voxel and the RGB pixel, there exists a strong correlation between adjacent voxels. This is analogous to the adjacent pixels' correlation in most of the images.
Spatial predictive coding is an efficient approach to exploiting statistical dependence between spatially adjacent samples. Therefore, we propose a spatial prediction method to reduce the information redundancy in the voxel grid.
\\
\textbf{Reference Graph Construction.}
In order to model the bit cost of residuals, we first construct a reference relation graph. For each remaining voxel, we select one adjacent preceding voxel as a reference, where preceding voxels represent voxels previously reconstructed in the decoding process. Although more refined spatial predictive coding may rely on interpolation to obtain feature values at non-integer positions as references, we only use complete voxels as prediction values for training convenience.

In this work, we use the Sum of Absolute Differences (SAD) as the metric to choose the optimal reference and record the reference index for each voxel in the bitstream, as shown in Fig.~\ref{refs}. This approach allows us to generate optimal prediction values for each voxel, albeit incurring a high bit cost of indexes. It is also feasible in our framework to partition voxels into blocks and select a mode for all voxels within a block, making a trade-off between prediction accuracy and index cost.

Compared to density grids, color feature grids typically cost much more storage. Therefore we only perform prediction on the feature grid for simplification.
\\
\textbf{Quantization and Prediction.}
In order to improve parallelism, we first quantize the feature vector and then predict instead of the traditional way, which performs quantization after prediction. Thus the residual can be calculated by,
\begin{equation}
\label{pred}
    y = \hat{x} - \hat{x}_{ref},
\end{equation}
where $\hat{x}$ and $\hat{x}_{ref}$ denotes the quantized feature to be coded and referred, respectively, and $y$ denotes the residual. We use a straightforward scalar quantization approach instead of vector quantization. Although vector quantization is typically more efficient, it may bring difficulties to subsequent prediction and bitrate estimation. Besides, we discover that the residual coefficients after finetuning are almost zero as shown in Fig. \ref{distribution}. The scalar quantization is efficient enough to handle such coefficients.

\subsection{Rate-Distortion Joint Finetuning}
Most current techniques for NeRF compression treat compression and 3D-scene representation as two independent optimization problems, which reduces their compression performance and limits the effectiveness of these methods in a wider bitrate range. In contrast, we train the NeRF model for bitrate and distortion simultaneously. We propose a simple entropy model to estimate the bitrate. Introducing a Lagrangian multiplier as a trade-off factor to balance the bitrate and distortion, the loss function is defined as follows,
\begin{equation}
    Loss = \lambda\times \mathcal{R}+\mathcal{D},
\end{equation}
where $\lambda$ is a hyperparameter, $\mathcal{R}$ is the bitrate shown in Eq. \ref{rate_def}, $\mathcal{D}$ is the distortion loss same as the loss function in the original method. We can obtain an optimal RD curve by simply adjusting the trade-off factor $\lambda$.
\\
\textbf{Quantization in Training Process.}
\begin{figure}[t]
    \centering
  \begin{subfigure}{0.49\linewidth}
    \includegraphics[width=\linewidth]{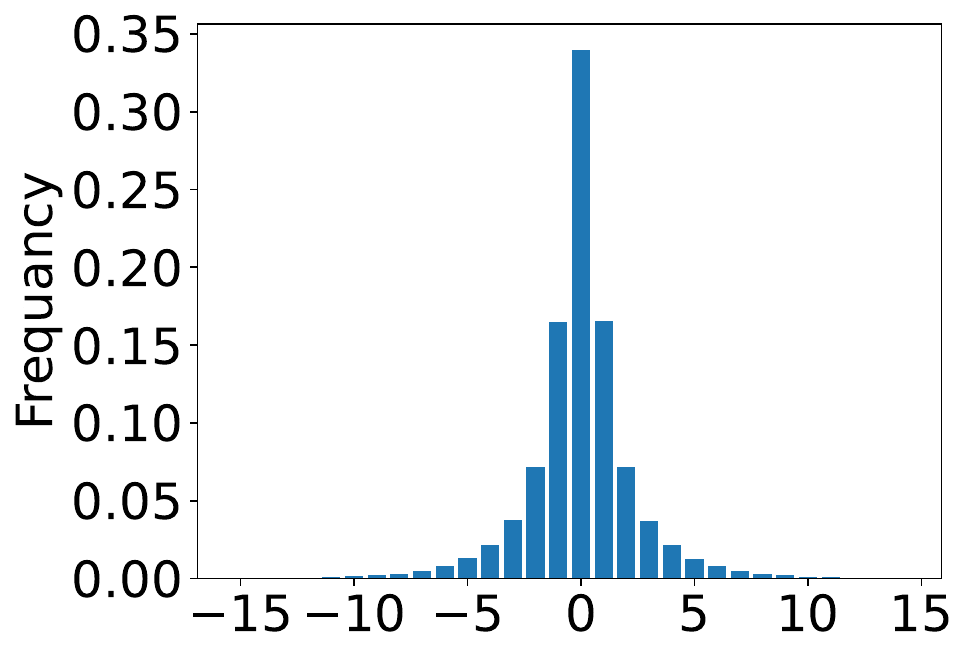} 
    \caption{Before finetuning}
  \end{subfigure}
  \hfill
  \begin{subfigure}{0.48\linewidth}
    \includegraphics[width=\linewidth]{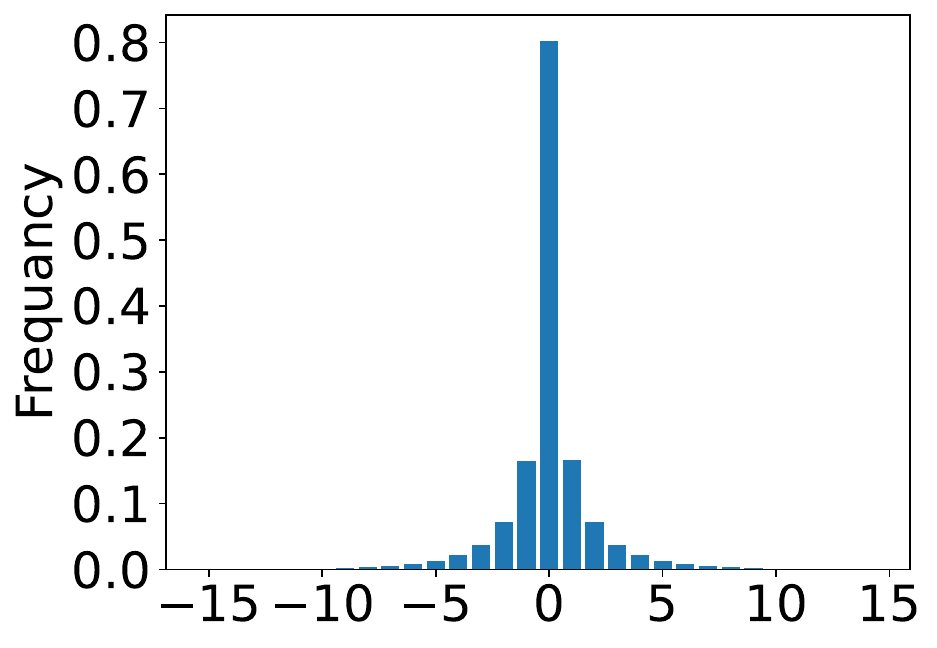} 
    \caption{After finetuning}
  \end{subfigure}
\caption{
Distribution prediction residuals in synthetic lego (quantization step = 0.5).
}
\label{distribution}
\end{figure}
The feature vector $x$ is designed to be scalar quantized with a uniform quantizer.
However, because the gradient of the quantization process equals zero almost everywhere, we adopt the method in~\cite{balle2018variational}, using uniform noise to substitute quantization error for backpropagation,
\begin{equation}
    \widetilde{x} = \frac{x}{\mathcal{Q}_{step}} + \mathcal{U}(-\frac{1}{2}, \frac{1}{2}),
\end{equation}
where $\widetilde{x}$ denotes the quantized feature while training, $\mathcal{Q}_{step}$ is the quantization step and $\mathcal{U}$ denotes the uniform distribution.
\\
\textbf{Entropy Modeling.} 
Considering the predictive coding paradigm in Fig. \ref{framework}, the reference indexes are fixed while training. Thus the bits for coding the indexes are constant. Because constant terms in the loss function have no effect on the training process and can be discarded, only the residual bits need to be modeled. Fig. \ref{distribution} shows the typical distribution of prediction residuals. We use an exponential distribution to simulate the distribution
and replace the bitrate with the entropy of residual coefficients,
\begin{equation}
    \mathcal{R} = \mathbb{E}_{y\sim EXP(\epsilon)}-log(p(y))=\kappa \mathbb{E}(|y|) + \beta,
\end{equation}
where $EXP$ denotes the exponential distribution, $\kappa$ and $b$ are constant terms. We further assume that the quantization error conforms to the uniform distribution, then,
\begin{equation}
    \mathbb{E}|y| = \mathbb{E}|\hat{x} - \hat{x}_{ref}| \approx \mathbb{E}|\widetilde{x} - \widetilde{x}_{ref}| = \frac{\mathbb{E}|x-x_{ref}|}{Q_{step}}.
\end{equation}
Since the constant term in the loss function has no contribution, and $\frac{\kappa}{Q_{step}}$ may be absorbed by the hyperparameter $\lambda$, the rate loss can be represented as a first-order regularizer for all residuals,
\begin{equation}
    \mathcal{R} = \frac{1}{|\mathcal{G}|} \sum_{(\bf{v}_i, \bf{v}_j)\in \mathcal{G}} \left\|\bf{v}_i-\bf{v}_j\right\|_1,
    \label{rate_def}
\end{equation}
where $|\mathcal{G}|$ is the number of edges in the reference graph $\mathcal{G}$, $\bf{v}$ denotes the feature voxel.

\subsection{Two-step Finetuning Procedure}
Although we compensate for the quantization loss with random noise during training, large quantization steps still lead to unacceptable performance loss, as shown in Table \ref{ablation}, while finer quantization costs more bits.
Considering that voxels are not equally important to the rendering results, we propose a two-step finetuning procedure combined with a hierarchical quantization strategy in this section, which performs precise quantization only for critical voxels to maximize bit saving without causing excessive rendering performance degradation.
\\
\textbf{Post Finetuning.}
A coarse feature grid is reconstructed after quantization and dequantization. Then we freeze all of the non-critical voxels in the feature grid and finetune the remaining parameters. Non-pruned volume density and the RGB network are still included in this stage to better compensate for quantization loss. A similar RD loss function is used as follows,
\begin{figure}[t]
\centering
\includegraphics[width=\columnwidth]{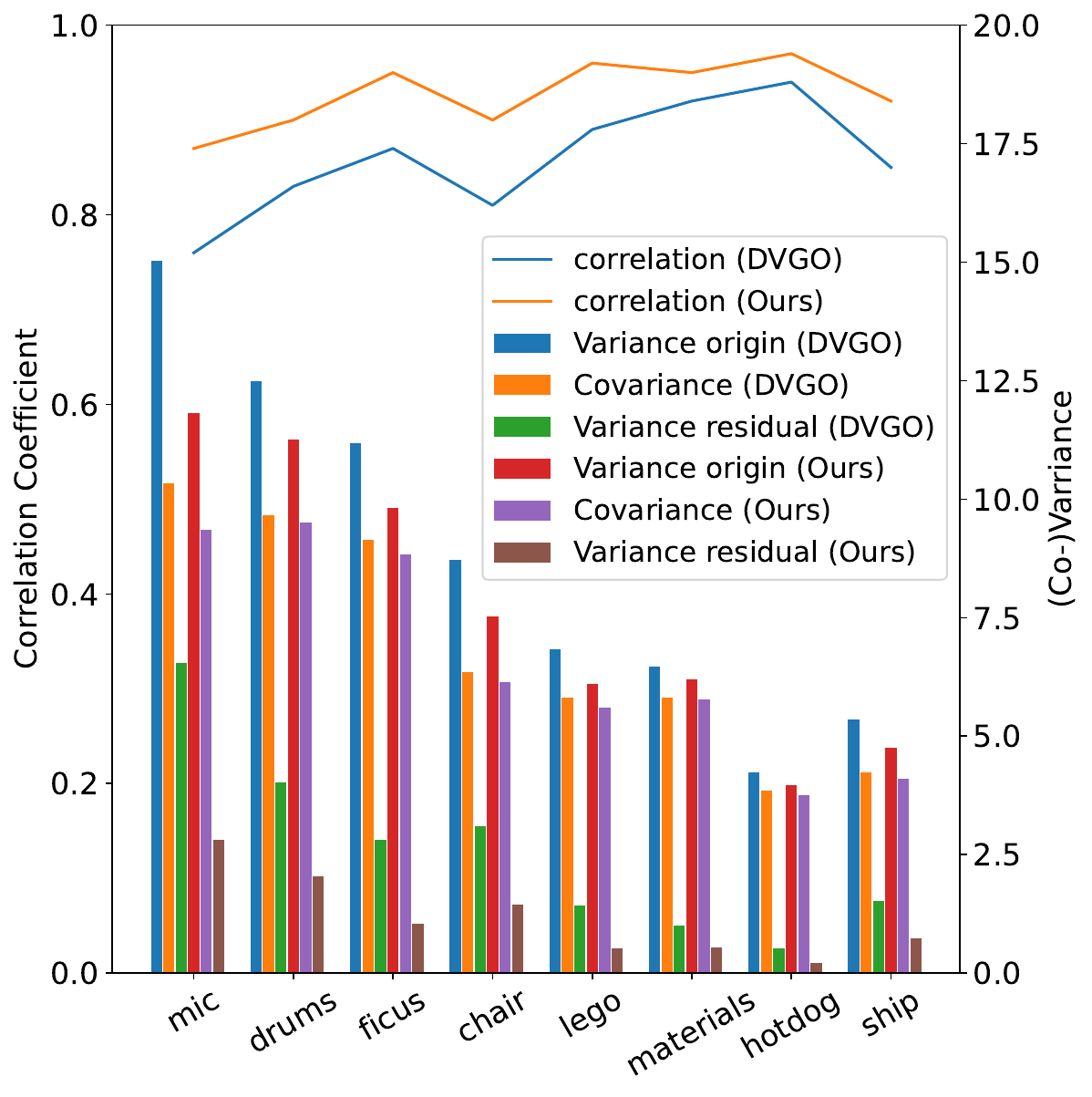} 
\caption{
Analysis of the spatial correlation between adjacent voxels in both uncompressed DVGO and our proposed SPC-NeRF. We calculate the covariance and the correlation coefficients between voxels and their reference voxels. the variances of original voxels and prediction residuals are also shown in this figure.
}
\label{correlation}
\end{figure}
\begin{equation}
    \mathcal{R} = \frac{1}{|\mathcal{C}|} \sum_{\bf{v}_i\in \mathcal{C}} \left\|\bf{v}_i-\hat{\bf{v}}_i\right\|_1,
\end{equation}
where $\mathcal{C}$ is the set of all critical voxels, and $\hat{\bf{v}}$ denotes the feature voxel in the coarse grid.

\begin{table*}[t]
\centering

\setlength{\tabcolsep}{5pt} 
\begin{tabular}{l|ccc|ccc|ccc|ccc}
\toprule
\multirow{3}{*}{Methods} &
\multicolumn{3}{c|}{Synthetic NeRF} &
\multicolumn{3}{c|}{Synthetic-NSVF} &
\multicolumn{3}{c|}{Blended-MVS} &
\multicolumn{3}{c}{T\&T}\\
& PSNR & SSIM & Size & PSNR & SSIM & Size & PSNR & SSIM & Size & PSNR & SSIM & Size
\\
& (dB)$\uparrow$ & $\uparrow$ & (MB)$\downarrow$ & (dB)$\uparrow$ &$\uparrow$ & (MB)$\downarrow$ & (dB)$\uparrow$ &$\uparrow$ & (MB)$\downarrow$ & (dB)$\uparrow$ &$\uparrow$ & (MB)$\downarrow$
\\
\midrule
\multicolumn{11}{l}{Non-voxel based NeRFs:} \\
\midrule
NeRF & 31.01 & 0.947 & 2.5 & - & - & - & - & - & - & 25.78 & 0.864 & 5.0 \\
NSVF & 31.74 & 0.953 & $\sim$16 & 35.13 & 0.979 & $\sim$16 & 26.89 & 0.898 & $\sim$16 & 28.40 & 0.900 & $\sim$16 \\
Instant-NGP & 33.04 & 0.934 & 28.64 & 36.11 & 0.966 & 46.09 & - & - & - & 28.81 & 0.917 & 46.09 \\
TensoRF & 33.14 & 0.963 & 71.8 & 36.52 & 0.982 & 71.8 & - & - & - & 28.56 & 0.920 & 71.8 \\
TensoRF-CP & 31.56 & 0.949 & 3.9 & 34.48 & 0.971 & 3.9 & - & - & - & 27.59 & 0.897 & 3.9 \\
\midrule
\multicolumn{11}{l}{Explicit voxel grid represented NeRFs:} \\
\midrule
Plenoxels &  31.71 & 0.958 & 259.8 & 34.12 & 0.977 & 283.3 & 26.84 & 0.911 & 367.7 & 26.84 & 0.911 & 367.7 \\
DVGO & 31.91 & 0.956 & 112.5 & 34.90 & 0.975 & 119.8 & 28.10 & 0.922 & 119.4 & 28.27 & 0.910 & 113.2 \\
Re:~NeRF & 31.08 & 0.944 & 2.0 & 34.90 & 0.969 & 2.46 & - & - & - & 27.90 & 0.894 & 1.62\\
TinyNeRF & 31.72 & 0.954 & 2.0 & 34.62 & 0.968 & 2.0 & 28.02 & 0.919 & 2.0 & 28.19 & 0.908 & 2.0 \\
VQRF & 31.77 & 0.954 & 1.431 & 34.63 & 0.974 & 1.263 & 27.93 & 0.918 & 1.529 & 28.18 & 0.909 & 1.417\\
Ours-high & 31.75 & 0.954 & 1.151 & 34.60 & 0.973 & 1.051 & 27.94 & 0.915 & 1.051 & 28.15 & 0.907 & 0.986\\
Ours-low & 31.50& 0.952& 0.819 & 34.31 & 0.970 &  0.745 & 27.65& 0.905& 0.811 & 27.93 & 0.903 & 0.775\\

\midrule
\multicolumn{11}{l}{EVG based NeRFs with higher voxel resolution:} \\
\midrule
DVGO-HR & 32.77 & 0.962 & 409.8 & 36.18 & 0.981 & 450.8 & 28.60 & 0.933 & 443.8 & 28.66 & 0.920 & 417.4 \\
VQRF-HR & 32.69 & 0.960 & 4.580 & 36.02 & 0.980 & 3.945 & 28.40 & 0.930 & 4.887 & 28.59 & 0.919 & 4.263 \\
Ours-HR-high & 32.70 & 0.961 & 3.771 & 36.07 & 0.980 & 3.343 & 28.52 & 0.930 & 3.639 & 28.63 & 0.919 & 3.295\\
Ours-HR-low & 32.59 & 0.960 & 2.851 & 35.90 & 0.979 & 2.534 & 28.39 & 0.925 & 2.768 & 28.52 & 0.916 & 2.550\\

\bottomrule
\end{tabular}
\caption{Comparisons on model size and rendering fidelity with state-of-the-art methods (Re:~NeRF and VQRF in the table refers to their DVGO-based implementation). We don't regard Instant-NGP~\cite{muller2022instant} and TensorRF~\cite{chen2022tensorf} as explicit voxel based NeRFs in this table. Although both of them utilize the concept of the voxel, Instant-NGP substitutes the explicit voxel grid with a hash table, while TensoRF decomposes the voxel grid into matrices or vectors. We conduct our experiment with the voxel resolution set to $160^3$ and $256^3$~(HR). We include a high-quality configuration and a low-quality configuration. The result shows our method takes minimal storage to obtain similar rendering quality to other methods. }
\label{main performance}
\end{table*}
\section{Experimental Results}
\subsection{Datasets}
We test our method on four common-used online datasets. All the scenes in these datasets contain a picture set for training and a picture set for testing.
\\
\textbf{Synthetic-NeRF}. Introduced by NeRF~\cite{10.1007/978-3-030-58452-8_24} and widely used in the following works, including 8 computer rendered scenes. Each scene contains 300 camera poses, 100 for training and 200 for testing.
\\
\textbf{Synthetic-NSVF} \cite{liu2020neural}. 8 more scenes rendered by the computer with more complex texture details and lighting conditions.
\\
\textbf{Blended-MVS} \cite{yao2020blendedmvs}. A synthetic multi-view stereo dataset with photo-level realistic ambient light. We use a subset provided by NSVF~\cite{liu2020neural}.
\\
\textbf{Tanks \& Temples} \cite{knapitsch2017tanks}. 1920$\times$1080 resolution pictures captured from the real world. We use 5 scenes~(Barn, Caterpillar, Family, Ignatius, Truch) for testing. 
\subsection{Implementation Details}
\begin{table}[t]
    \centering
    \begin{tabular}{l|cc}
        \toprule
         \multirow{2}{*}{Scenes} & \multicolumn{2}{c}{BD-Rate} \\
         & \ \ \ SPC-NeRF \ \ \  & SPC-NeRF-HR\\
         \midrule
         chair & -19.50\% & -25.86\% \\
         drums & -34.82\% & -19.18\%\\
         ficus & -15.28\% & -34.13\%\\
         hotdog & -24.09\% & -33.95\%\\
         lego & -27.52\% & -28.96\%\\
         materials & -47.43\% & -57.93\%\\
         mic & -22.90\% & -27.85\%\\
         ship & -35.68\% & -29.51\%\\
         \midrule
         Average & -28.40\% & -32.17\%\\
         \bottomrule
    \end{tabular}
    \caption{Piecewise cubic BD-Rate (PSNR) performance on Synthetic-NeRF dataset. Bit saving percentages with same PSNR compared to VQRF (anchor).}
    \label{bd-rate}
\end{table}
We apply the pipeline in DVGO\footnote{\url{https://github.com/sunset1995/DirectVoxGO}} and compress the model generated by DVGO. We use a pruning process similar to VQRF\footnote{\url{https://github.com/AlgoHunt/VQRF}} with the pruning and critical quantiles set to 0.001 and 0.6, respectively. In our fintuneing process, the learning rates are initialized to be 0.1 for all voxel grids and to be 0.001 for the network weights. Furthermore, the learning rates exponentially decay during training. We do not reinitialize the learning rates before post finetuning, i.e., learning rates at the beginning of post finetuning are identical to learning rates at the end of the previous finetuning process. We test a base configuration and a Higher Resolution~(HR) configuration for both our method and VQRF, wherein the expected numbers of voxels are set to $160^3$ and $256^3$, respectively. Please refer to the supplementary material for more detailed training configuration.

The syntax elements to reconstruct the feature grid (coarse residuals, fine residuals and reference indexes) are entropy coded via a sample arithmetic coding implementation\footnote{\url{https://github.com/nayuki/Reference-arithmetic-coding}}.
Specifically, we rearrange the reference candidates by putting the most recently selected candidate in the first position to further improve the entropy coding efficiency. Other syntax elements that take up little storage are coded with an LZ77 codec~\cite{ziv1977universal}. We conduct our experiments on an Nvidia RTX 4090 graphics card with Cuda version 11.8. 

\subsection{Performance Comparisons}
\textbf{Quantitative Results.}
We compare our method with non-voxel based NeRFs~\cite{chen2022tensorf, muller2022instant, 10.1007/978-3-030-58452-8_24, liu2020neural},
EVG-based NeRFs~\cite{sun2022direct, fridovich2022plenoxels}, and recent compressed EVG NeRFs~\cite{deng2023compressing, li2023compressing, zhao2023tinynerf}, which represent the state-of-the-art performance. We adjust the hyperparameter $\lambda$ to make the rendering quality of our method as close as possible to the SOTA method. The results in Table \ref{main performance} show that our method achieves about 100$\times$ compression on the uncompressed DVGO, with a slight loss of rendering quality. In comparison with other compression methods for EVG-based NeRFs, our method has the smallest model size under the same rendering quality. 

We also test the performance of our method with higher compression ratios by adjusting the trade-off factor $\lambda$. We display the result of a low-quality configuration in Table \ref{main performance} where $\lambda$ is four times larger. It suggests that our method uses nearly half the model size to achieve similar rendering quality compared to VQRF, where the average PSNR performance drop merely 0.3 dB.

For a fair comparison of the overall rate-distortion performance, we introduce the piecewise cubic Bjøntegaard-Delta-Rate (BD-Rate) metric~\cite{bjontegaard2001calculation}. The metric employs piecewise cubic interpolation to fit the continuous RD curve and calculate the bit saving at the same quality. We adjust the pruning quantiles in VQRF to obtain the results of different compression ratios. Table \ref{bd-rate} shows the BD-Rate results on the Synthetic-NeRF dataset. our method achieves 28\% bit saving for the base configuration and 32\% bit saving for the HR configuration.

\begin{figure}[t]
\centering
  \begin{subfigure}{0.486\linewidth}
    \includegraphics[width=\linewidth]{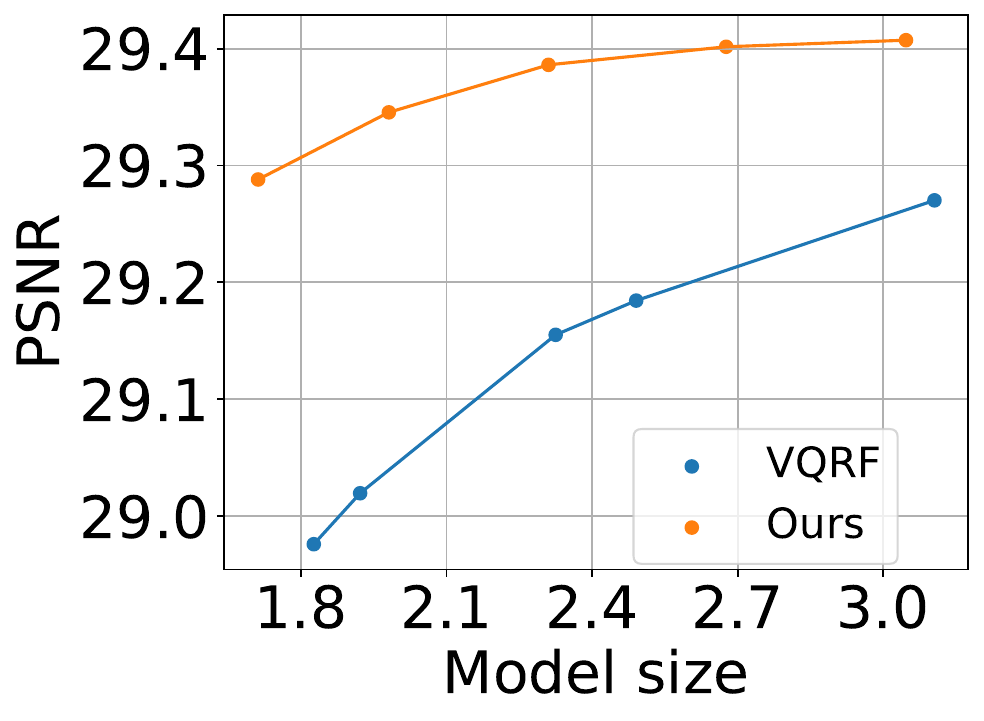} 
    \caption{Materials}
  \end{subfigure}
  \hfill
  \begin{subfigure}{0.49\linewidth}
    \includegraphics[width=\linewidth]{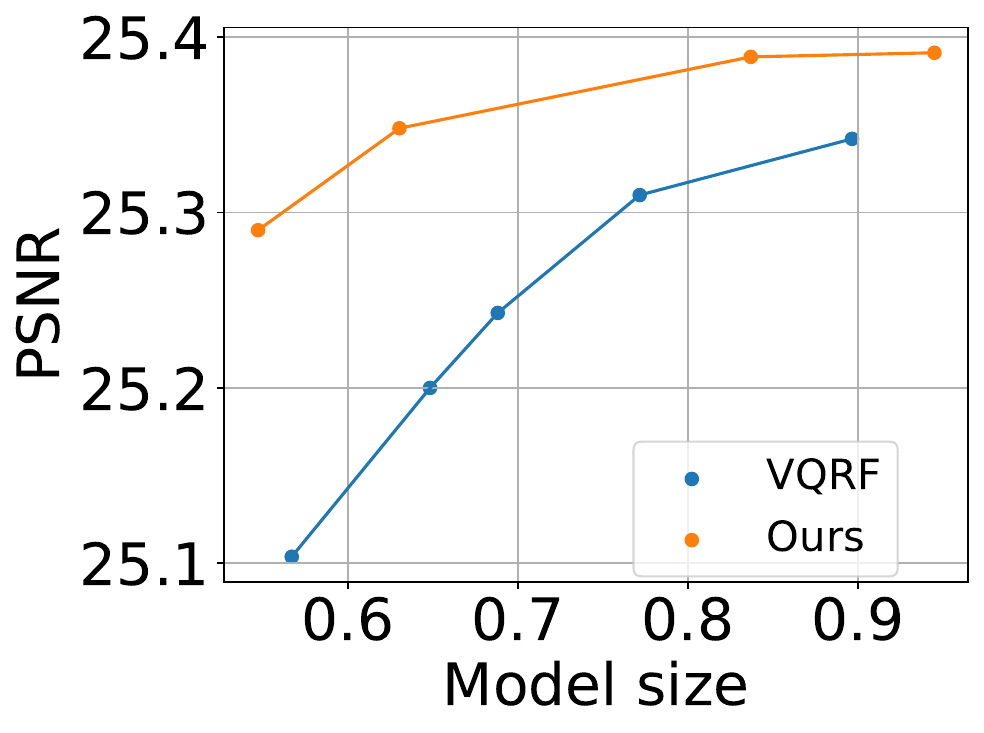} 
    \caption{Drums}
  \end{subfigure}
  
  \begin{subfigure}{0.493\linewidth}
    \includegraphics[width=\linewidth]{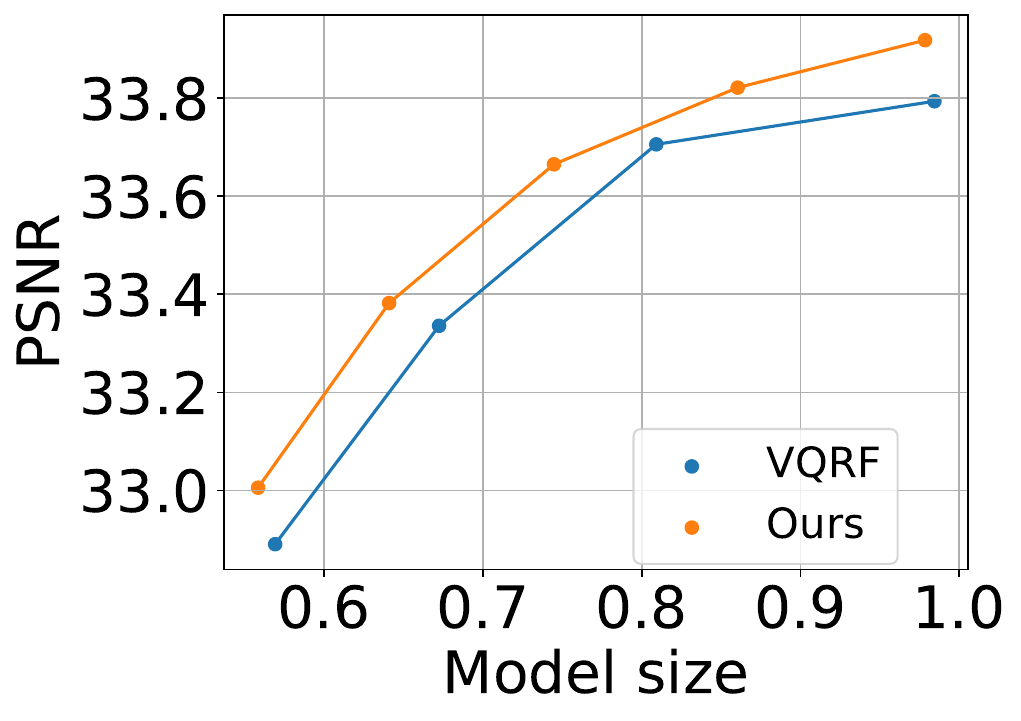} 
    \caption{Chair}
  \end{subfigure}
  \hfill
  \begin{subfigure}{0.48\linewidth}
    \includegraphics[width=\linewidth]{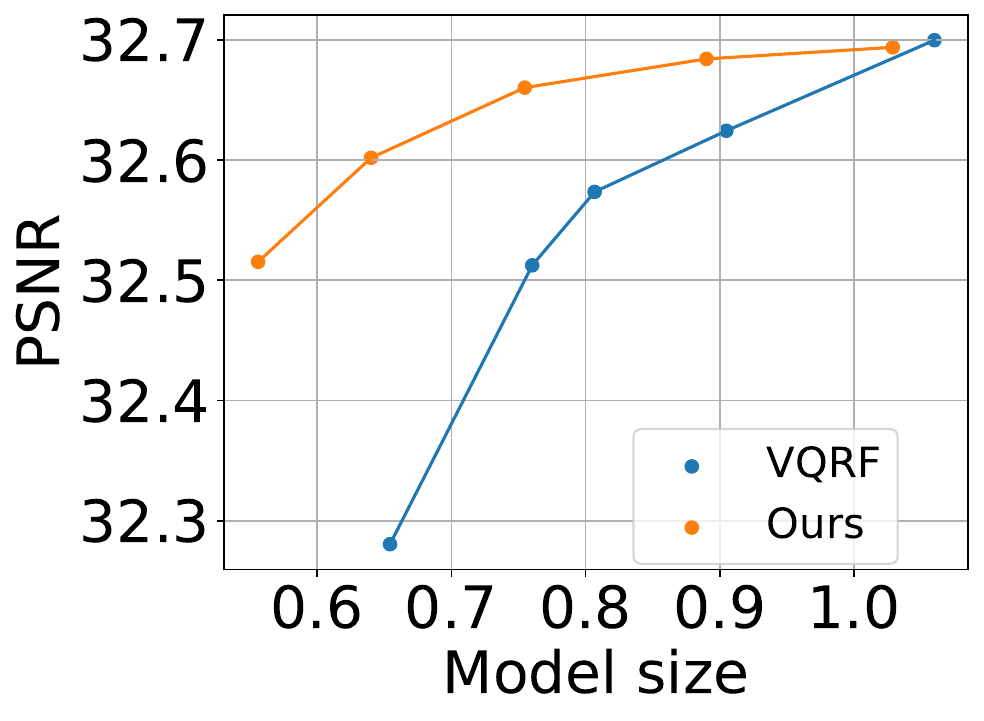} 
    \caption{Ficus}
  \end{subfigure}
\caption{
    RD curves of four typical scenes. The x-axis represents model size (MB) and the y-axis represents PSNR (dB). The VQRF curves are not smooth because the different model sizes are obtained by adjusting pruning strength.
}
\label{scene-rd-curve}
\end{figure}

We also generate the RD curves of four typical scenes (Fig. \ref{scene-rd-curve}) for more intuitive comparisons. The average RD curve can also be seen in Fig. \ref{main performance}. The results reveal that VQRF works relatively better on high bitrates, but the rendering quality drops sharply as the bitrate decreases, because either increasing the pruning strength or quantizing the critical voxels results in rapid losses of texture details on the edges. Besides, the VQRF performs worse in reconstructing scenes with abundant details because the limited codebook has difficulties handling evenly distributed feature vectors. In contrast, our SPC-NeRF can generate a smooth approximate logarithmic RD curve by simply adjusting $\lambda$.

\begin{figure*}[t]
\centering
\includegraphics[width=0.95\textwidth]{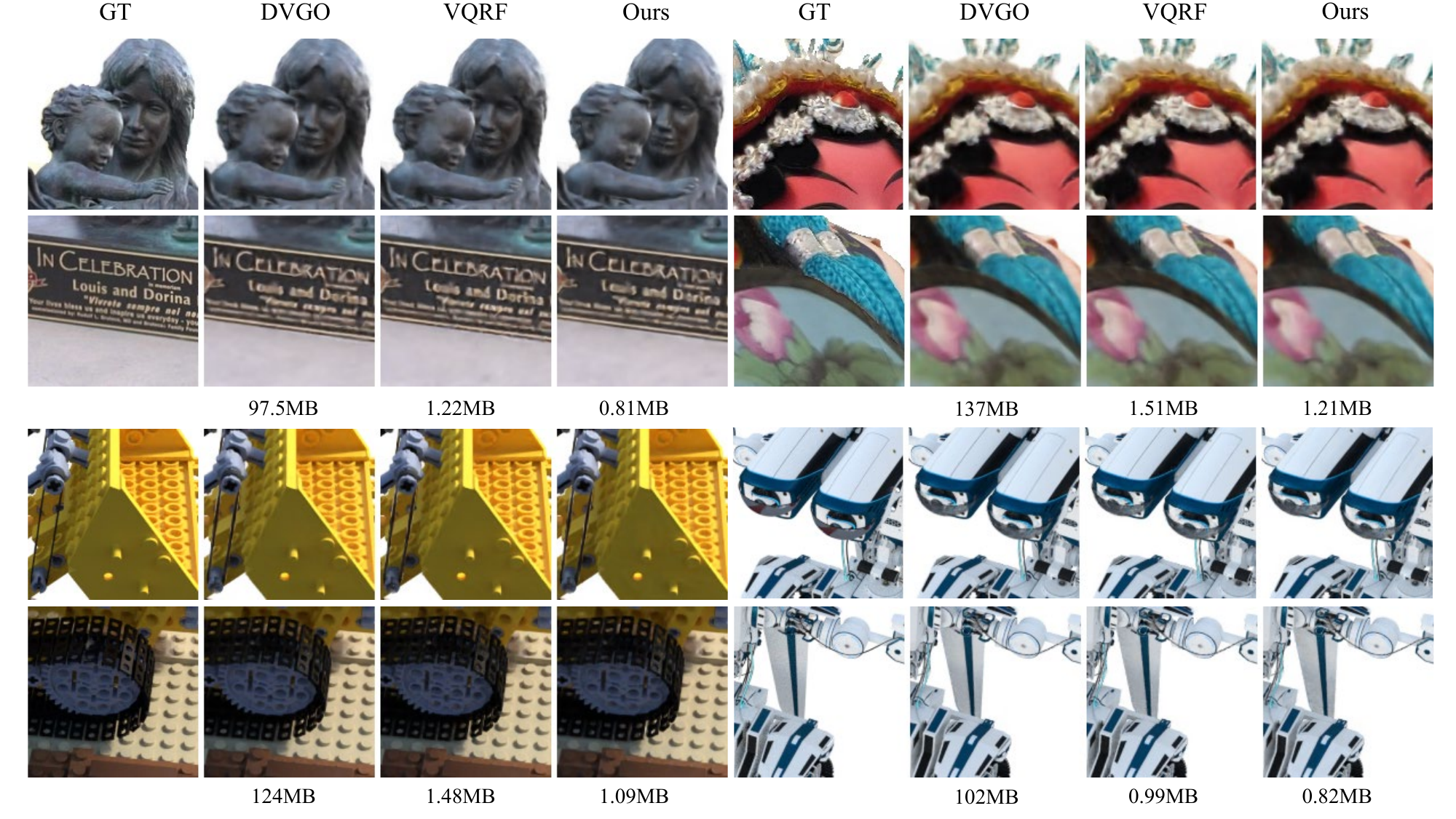} 
\caption{Subjective comparison. We selected four challenging scenes from different datasets, Family, Character, Lego, and Robot, arranged from top to bottom and left to right. Subjective quality degradation of our method compared to the uncompressed DVGO can hardly be observed.}
\label{subjective result}
\end{figure*}

\textbf{Subjective Results.} We compare the rendering quality of our method with the uncompressed DVGO and VQRF on multiple challenging scenes. The results in Fig. \ref{subjective result} show that our method achieves comparable subjective quality to DVGO and VQ-DVGO with smaller model sizes.

\textbf{Speed Comparison.}
\begin{table}[t]
    \centering
    \begin{tabular}{l|ccc}
        \toprule
         Method & 
         Train~(min) & Render~(sec) \\
         \midrule
         DVGO & 3 & 36  \\
         VQRF & 3+2 & 38 \\
         Ours & 3+3 & 36 \\
         \bottomrule
    \end{tabular}
    \caption{Comparison of training and rendering time. We render 200 pictures of resolution $800\times 800$ and report the total time.}
    \label{training time}
\end{table}
Table \ref{training time} shows the time analysis of our methods. Our SPC-NeRF takes a little more training time than VQRF, and the rendering speed is comparable to other EVG NeRFs.

\subsection{Ablation Study}
We study the effect of the three proposed methods by switching them off separately. The results are shown in Table \ref{ablation}. The post-finetune process can use a small number of bits to enhance the reconstruction quality of important voxels, compensating for the losses introduced by coarse quantization. Although disabling post-finetune can save bits, it inevitably deteriorates the rendering quality.
The rate loss guides the training process to reduce the bitrate. Therefore a larger value of $\lambda$ in Eq.~\ref{rate_def} means the bitrate is more emphasized during training. Training without the rate loss is equivalent to setting $\lambda$ to 0, which leads to better rendering quality, accompanied by a rapid increase in the bitrate.
As the primary contribution of our work, spatial predictive coding plays a pivotal role in the compression efficiency of our approach. Therefore, we perform a more detailed analysis of the spatial predictive coding.

We analyse the spatial correlation between voxels and their reference voxels by calculating the correlation coefficient as shown in Fig.~\ref{correlation}. We also depict the variances of the the residuals and original features, respectively. The much lower energy of residuals indicates our spatial prediction method can efficiently remove the spatial redundancy.
Additionally, the statistics after finetuning compared with those in the original DVGO confirms the efficacy of our finetuning approach. Incorporating the bitrate of prediction residuals into the loss function and jointly training for rate and distortion can enhance the spatial correlation of the voxel grid. Consequently, it leads to a substantial reduction of the model size. Although the prediction does not contribute to the rendering quality directly, training and coding the voxel grid without prediction will still harm the overall RD performance. The main reason is that the RD loss function makes a balance between the model size and the rendering quality.


\begin{table}[t]
    \centering
    \begin{tabular}{l|ccc}
        \toprule
         \multirow{2}{*}{Method} & 
         $\Delta$PSNR & $\Delta$SSIM & $\Delta$ Size\\
         & (dB)$\uparrow$ & $\uparrow$ & (MB)$\downarrow$\\
         \midrule
         w/o prediction & -0.09 & -0.001 & +0.669  \\
         w/o post finetune & -1.72 & -0.07 & -0.079 \\
         w/o rate loss & +0.13 & +0.002 & +0.633 \\
         \bottomrule
    \end{tabular}
    \caption{Ablation study on Synthetic-NeRF dataset.}
    \label{ablation}
\end{table}


\section{Conclusion}
In this paper, we performed a detailed analysis of the spatial correlation in the radiance field represented explicitly by the voxel grid and experimentally showed there exists substantial redundancy. Aiming at reducing such redundancy, we proposed SPC-NeRF, a novel spatial predictive coding framework, and designed a finetuning procedure to generate spatially regularized radiance field.
We tested the proposed method with two different configurations on four representative datasets. The experimental results showed that the proposed method can achieve 32\% bit rate saving, compared with the state-of-the-art method VQRF.

{
    \small
    \bibliographystyle{ieeenat_fullname}
    \bibliography{main}

\begin{thebibliography}{46}
\providecommand{\natexlab}[1]{#1}
\providecommand{\url}[1]{\texttt{#1}}
\expandafter\ifx\csname urlstyle\endcsname\relax
  \providecommand{\doi}[1]{doi: #1}\else
  \providecommand{\doi}{doi: \begingroup \urlstyle{rm}\Url}\fi

\bibitem[Ball{\'e} et~al.(2018)Ball{\'e}, Minnen, Singh, Hwang, and
  Johnston]{balle2018variational}
Johannes Ball{\'e}, David Minnen, Saurabh Singh, Sung~Jin Hwang, and Nick
  Johnston.
\newblock Variational image compression with a scale hyperprior.
\newblock \emph{arXiv preprint arXiv:1802.01436}, 2018.

\bibitem[Bao et~al.(2023)Bao, Zhang, Yang, Fan, Yang, Bao, Zhang, and
  Cui]{bao2023sine}
Chong Bao, Yinda Zhang, Bangbang Yang, Tianxing Fan, Zesong Yang, Hujun Bao,
  Guofeng Zhang, and Zhaopeng Cui.
\newblock Sine: Semantic-driven image-based nerf editing with prior-guided
  editing field.
\newblock In \emph{Proceedings of the IEEE/CVF Conference on Computer Vision
  and Pattern Recognition}, pages 20919--20929, 2023.

\bibitem[Bird et~al.(2021)Bird, Ball{\'e}, Singh, and Chou]{bird20213d}
Thomas Bird, Johannes Ball{\'e}, Saurabh Singh, and Philip~A Chou.
\newblock 3d scene compression through entropy penalized neural representation
  functions.
\newblock In \emph{2021 Picture Coding Symposium (PCS)}, pages 1--5. IEEE,
  2021.

\bibitem[Bjontegaard(2001)]{bjontegaard2001calculation}
Gisle Bjontegaard.
\newblock Calculation of average psnr differences between rd-curves.
\newblock \emph{ITU SG16 Doc. VCEG-M33}, 2001.

\bibitem[Bross et~al.(2014)Bross, Helle, Lakshman, and Ugur]{bross2014inter}
Benjamin Bross, Philipp Helle, Haricharan Lakshman, and Kemal Ugur.
\newblock Inter-picture prediction in hevc.
\newblock \emph{High Efficiency Video Coding (HEVC) Algorithms and
  Architectures}, pages 113--140, 2014.

\bibitem[Bross et~al.(2021)Bross, Wang, Ye, Liu, Chen, Sullivan, and
  Ohm]{bross2021overview}
Benjamin Bross, Ye-Kui Wang, Yan Ye, Shan Liu, Jianle Chen, Gary~J Sullivan,
  and Jens-Rainer Ohm.
\newblock Overview of the versatile video coding (vvc) standard and its
  applications.
\newblock \emph{IEEE Transactions on Circuits and Systems for Video
  Technology}, 31\penalty0 (10):\penalty0 3736--3764, 2021.

\bibitem[Cao and Johnson(2023)]{cao2023hexplane}
Ang Cao and Justin Johnson.
\newblock Hexplane: A fast representation for dynamic scenes.
\newblock In \emph{Proceedings of the IEEE/CVF Conference on Computer Vision
  and Pattern Recognition}, pages 130--141, 2023.

\bibitem[Chen et~al.(2022)Chen, Xu, Geiger, Yu, and Su]{chen2022tensorf}
Anpei Chen, Zexiang Xu, Andreas Geiger, Jingyi Yu, and Hao Su.
\newblock Tensorf: Tensorial radiance fields.
\newblock In \emph{Computer Vision--ECCV 2022: 17th European Conference, Tel
  Aviv, Israel, October 23--27, 2022, Proceedings, Part XXXII}, pages 333--350.
  Springer, 2022.

\bibitem[Choi et~al.(2018)Choi, Wang, Venkataramani, Chuang, Srinivasan, and
  Gopalakrishnan]{choi2018pact}
Jungwook Choi, Zhuo Wang, Swagath Venkataramani, Pierce I-Jen Chuang,
  Vijayalakshmi Srinivasan, and Kailash Gopalakrishnan.
\newblock Pact: Parameterized clipping activation for quantized neural
  networks.
\newblock \emph{arXiv preprint arXiv:1805.06085}, 2018.

\bibitem[Deng and Tartaglione(2023)]{deng2023compressing}
Chenxi~Lola Deng and Enzo Tartaglione.
\newblock Compressing explicit voxel grid representations: fast nerfs become
  also small.
\newblock In \emph{Proceedings of the IEEE/CVF Winter Conference on
  Applications of Computer Vision}, pages 1236--1245, 2023.

\bibitem[Duan et~al.(2022)Duan, Liu, Jia, Wang, Ma, and
  Gao]{duan2022differential}
Wenhong Duan, Zhenhua Liu, Chuanmin Jia, Shanshe Wang, Siwei Ma, and Wen Gao.
\newblock Differential weight quantization for multi-model compression.
\newblock \emph{IEEE Transactions on Multimedia}, 2022.

\bibitem[Fang et~al.(2022)Fang, Yi, Wang, Xie, Zhang, Liu, Nie{\ss}ner, and
  Tian]{fang2022fast}
Jiemin Fang, Taoran Yi, Xinggang Wang, Lingxi Xie, Xiaopeng Zhang, Wenyu Liu,
  Matthias Nie{\ss}ner, and Qi Tian.
\newblock Fast dynamic radiance fields with time-aware neural voxels.
\newblock In \emph{SIGGRAPH Asia 2022 Conference Papers}, pages 1--9, 2022.

\bibitem[Fridovich-Keil et~al.(2022)Fridovich-Keil, Yu, Tancik, Chen, Recht,
  and Kanazawa]{fridovich2022plenoxels}
Sara Fridovich-Keil, Alex Yu, Matthew Tancik, Qinhong Chen, Benjamin Recht, and
  Angjoo Kanazawa.
\newblock Plenoxels: Radiance fields without neural networks.
\newblock In \emph{Proceedings of the IEEE/CVF Conference on Computer Vision
  and Pattern Recognition}, pages 5501--5510, 2022.

\bibitem[Fridovich-Keil et~al.(2023)Fridovich-Keil, Meanti, Warburg, Recht, and
  Kanazawa]{fridovich2023k}
Sara Fridovich-Keil, Giacomo Meanti, Frederik~Rahb{\ae}k Warburg, Benjamin
  Recht, and Angjoo Kanazawa.
\newblock K-planes: Explicit radiance fields in space, time, and appearance.
\newblock In \emph{Proceedings of the IEEE/CVF Conference on Computer Vision
  and Pattern Recognition}, pages 12479--12488, 2023.

\bibitem[Gordon et~al.(2023)Gordon, Chng, MacDonald, and
  Lucey]{Gordon_2023_WACV}
Cameron Gordon, Shin-Fang Chng, Lachlan MacDonald, and Simon Lucey.
\newblock On quantizing implicit neural representations.
\newblock In \emph{Proceedings of the IEEE/CVF Winter Conference on
  Applications of Computer Vision (WACV)}, pages 341--350, 2023.

\bibitem[Gou et~al.(2021)Gou, Yu, Maybank, and Tao]{gou2021knowledge}
Jianping Gou, Baosheng Yu, Stephen~J Maybank, and Dacheng Tao.
\newblock Knowledge distillation: A survey.
\newblock \emph{International Journal of Computer Vision}, 129:\penalty0
  1789--1819, 2021.

\bibitem[Han et~al.(2021)Han, Li, Mukherjee, Chiang, Grange, Chen, Su, Parker,
  Deng, Joshi, et~al.]{han2021technical}
Jingning Han, Bohan Li, Debargha Mukherjee, Ching-Han Chiang, Adrian Grange,
  Cheng Chen, Hui Su, Sarah Parker, Sai Deng, Urvang Joshi, et~al.
\newblock A technical overview of av1.
\newblock \emph{Proceedings of the IEEE}, 109\penalty0 (9):\penalty0
  1435--1462, 2021.

\bibitem[He et~al.(2017)He, Zhang, and Sun]{he2017channel}
Yihui He, Xiangyu Zhang, and Jian Sun.
\newblock Channel pruning for accelerating very deep neural networks.
\newblock In \emph{Proceedings of the IEEE international conference on computer
  vision}, pages 1389--1397, 2017.

\bibitem[Knapitsch et~al.(2017)Knapitsch, Park, Zhou, and
  Koltun]{knapitsch2017tanks}
Arno Knapitsch, Jaesik Park, Qian-Yi Zhou, and Vladlen Koltun.
\newblock Tanks and temples: Benchmarking large-scale scene reconstruction.
\newblock \emph{ACM Transactions on Graphics (ToG)}, 36\penalty0 (4):\penalty0
  1--13, 2017.

\bibitem[Lei et~al.(2020)Lei, Luo, Zhang, Wang, and Ma]{9190915}
Meng Lei, Falei Luo, Xinfeng Zhang, Shanshe Wang, and Siwei Ma.
\newblock Two-step progressive intra prediction for versatile video coding.
\newblock In \emph{2020 IEEE International Conference on Image Processing
  (ICIP)}, pages 1137--1141, 2020.

\bibitem[Li et~al.(2016)Li, Kadav, Durdanovic, Samet, and Graf]{li2016pruning}
Hao Li, Asim Kadav, Igor Durdanovic, Hanan Samet, and Hans~Peter Graf.
\newblock Pruning filters for efficient convnets.
\newblock \emph{arXiv preprint arXiv:1608.08710}, 2016.

\bibitem[Li et~al.(2023)Li, Shen, Wang, Shen, and Bo]{li2023compressing}
Lingzhi Li, Zhen Shen, Zhongshu Wang, Li Shen, and Liefeng Bo.
\newblock Compressing volumetric radiance fields to 1 mb.
\newblock In \emph{Proceedings of the IEEE/CVF Conference on Computer Vision
  and Pattern Recognition}, pages 4222--4231, 2023.

\bibitem[Li et~al.(2022)Li, Slavcheva, Zollhoefer, Green, Lassner, Kim,
  Schmidt, Lovegrove, Goesele, Newcombe, et~al.]{li2022neural}
Tianye Li, Mira Slavcheva, Michael Zollhoefer, Simon Green, Christoph Lassner,
  Changil Kim, Tanner Schmidt, Steven Lovegrove, Michael Goesele, Richard
  Newcombe, et~al.
\newblock Neural 3d video synthesis from multi-view video.
\newblock In \emph{Proceedings of the IEEE/CVF Conference on Computer Vision
  and Pattern Recognition}, pages 5521--5531, 2022.

\bibitem[Lin et~al.(2019)Lin, Ji, Chen, Tao, and Luo]{8478366}
Shaohui Lin, Rongrong Ji, Chao Chen, Dacheng Tao, and Jiebo Luo.
\newblock Holistic cnn compression via low-rank decomposition with knowledge
  transfer.
\newblock \emph{IEEE Transactions on Pattern Analysis and Machine
  Intelligence}, 41\penalty0 (12):\penalty0 2889--2905, 2019.

\bibitem[Liu et~al.(2020)Liu, Gu, Zaw~Lin, Chua, and Theobalt]{liu2020neural}
Lingjie Liu, Jiatao Gu, Kyaw Zaw~Lin, Tat-Seng Chua, and Christian Theobalt.
\newblock Neural sparse voxel fields.
\newblock \emph{Advances in Neural Information Processing Systems},
  33:\penalty0 15651--15663, 2020.

\bibitem[Ma et~al.(2021)Ma, Lin, Ye, He, Zhang, Yuan, Tan, Li, Fan, Qian,
  et~al.]{ma2021non}
Xiaolong Ma, Sheng Lin, Shaokai Ye, Zhezhi He, Linfeng Zhang, Geng Yuan,
  Sia~Huat Tan, Zhengang Li, Deliang Fan, Xuehai Qian, et~al.
\newblock Non-structured dnn weight pruning—is it beneficial in any platform?
\newblock \emph{IEEE transactions on neural networks and learning systems},
  33\penalty0 (9):\penalty0 4930--4944, 2021.

\bibitem[Max(1995)]{max1995optical}
Nelson Max.
\newblock Optical models for direct volume rendering.
\newblock \emph{IEEE Transactions on Visualization and Computer Graphics},
  1\penalty0 (2):\penalty0 99--108, 1995.

\bibitem[Mildenhall et~al.(2020)Mildenhall, Srinivasan, Tancik, Barron,
  Ramamoorthi, and Ng]{10.1007/978-3-030-58452-8_24}
Ben Mildenhall, Pratul~P. Srinivasan, Matthew Tancik, Jonathan~T. Barron, Ravi
  Ramamoorthi, and Ren Ng.
\newblock Nerf: Representing scenes as neural radiance fields for view
  synthesis.
\newblock In \emph{Computer Vision -- ECCV 2020}, pages 405--421, Cham, 2020.
  Springer International Publishing.

\bibitem[M{\"u}ller et~al.(2022)M{\"u}ller, Evans, Schied, and
  Keller]{muller2022instant}
Thomas M{\"u}ller, Alex Evans, Christoph Schied, and Alexander Keller.
\newblock Instant neural graphics primitives with a multiresolution hash
  encoding.
\newblock \emph{ACM Transactions on Graphics (ToG)}, 41\penalty0 (4):\penalty0
  1--15, 2022.

\bibitem[Pfaff et~al.(2021)Pfaff, Filippov, Liu, Zhao, Chen,
  De-Lux{\'a}n-Hern{\'a}ndez, Wiegand, Rufitskiy, Ramasubramonian, and Van~der
  Auwera]{pfaff2021intra}
Jonathan Pfaff, Alexey Filippov, Shan Liu, Xin Zhao, Jianle Chen, Santiago
  De-Lux{\'a}n-Hern{\'a}ndez, Thomas Wiegand, Vasily Rufitskiy, Adarsh~Krishnan
  Ramasubramonian, and Geert Van~der Auwera.
\newblock Intra prediction and mode coding in vvc.
\newblock \emph{IEEE Transactions on Circuits and Systems for Video
  Technology}, 31\penalty0 (10):\penalty0 3834--3847, 2021.

\bibitem[Pumarola et~al.(2021)Pumarola, Corona, Pons-Moll, and
  Moreno-Noguer]{pumarola2021d}
Albert Pumarola, Enric Corona, Gerard Pons-Moll, and Francesc Moreno-Noguer.
\newblock D-nerf: Neural radiance fields for dynamic scenes.
\newblock In \emph{Proceedings of the IEEE/CVF Conference on Computer Vision
  and Pattern Recognition}, pages 10318--10327, 2021.

\bibitem[Rho et~al.(2023)Rho, Lee, Nam, Lee, Ko, and Park]{rho2023masked}
Daniel Rho, Byeonghyeon Lee, Seungtae Nam, Joo~Chan Lee, Jong~Hwan Ko, and
  Eunbyung Park.
\newblock Masked wavelet representation for compact neural radiance fields.
\newblock In \emph{Proceedings of the IEEE/CVF Conference on Computer Vision
  and Pattern Recognition}, pages 20680--20690, 2023.

\bibitem[Shi and Guillemot(2023)]{10095668}
Jinglei Shi and Christine Guillemot.
\newblock Light field compression via compact neural scene representation.
\newblock In \emph{ICASSP 2023 - 2023 IEEE International Conference on
  Acoustics, Speech and Signal Processing (ICASSP)}, pages 1--5, 2023.

\bibitem[Sun et~al.(2022)Sun, Sun, and Chen]{sun2022direct}
Cheng Sun, Min Sun, and Hwann-Tzong Chen.
\newblock Direct voxel grid optimization: Super-fast convergence for radiance
  fields reconstruction.
\newblock In \emph{Proceedings of the IEEE/CVF Conference on Computer Vision
  and Pattern Recognition}, pages 5459--5469, 2022.

\bibitem[Swaminathan et~al.(2020)Swaminathan, Garg, Kannan, and
  Andres]{swaminathan2020sparse}
Sridhar Swaminathan, Deepak Garg, Rajkumar Kannan, and Frederic Andres.
\newblock Sparse low rank factorization for deep neural network compression.
\newblock \emph{Neurocomputing}, 398:\penalty0 185--196, 2020.

\bibitem[Tang et~al.(2022)Tang, Chen, Wang, and Zeng]{tang2022compressible}
Jiaxiang Tang, Xiaokang Chen, Jingbo Wang, and Gang Zeng.
\newblock Compressible-composable nerf via rank-residual decomposition.
\newblock \emph{Advances in Neural Information Processing Systems},
  35:\penalty0 14798--14809, 2022.

\bibitem[Wang et~al.(2023)Wang, Hu, He, Wang, Yu, Tuytelaars, Xu, and
  Wu]{wang2023neural}
Liao Wang, Qiang Hu, Qihan He, Ziyu Wang, Jingyi Yu, Tinne Tuytelaars, Lan Xu,
  and Minye Wu.
\newblock Neural residual radiance fields for streamably free-viewpoint videos.
\newblock In \emph{Proceedings of the IEEE/CVF Conference on Computer Vision
  and Pattern Recognition}, pages 76--87, 2023.

\bibitem[Weinberger et~al.(2000)Weinberger, Seroussi, and Sapiro]{855427}
M.J. Weinberger, G. Seroussi, and G. Sapiro.
\newblock The loco-i lossless image compression algorithm: principles and
  standardization into jpeg-ls.
\newblock \emph{IEEE Transactions on Image Processing}, 9\penalty0
  (8):\penalty0 1309--1324, 2000.

\bibitem[Wiegand et~al.(2011)Wiegand, Schwarz, et~al.]{wiegand2011source}
Thomas Wiegand, Heiko Schwarz, et~al.
\newblock Source coding: Part i of fundamentals of source and video coding.
\newblock \emph{Foundations and Trends{\textregistered} in Signal Processing},
  4\penalty0 (1--2):\penalty0 151--179, 2011.

\bibitem[Witten et~al.(1987)Witten, Neal, and Cleary]{witten1987arithmetic}
Ian~H Witten, Radford~M Neal, and John~G Cleary.
\newblock Arithmetic coding for data compression.
\newblock \emph{Communications of the ACM}, 30\penalty0 (6):\penalty0 520--540,
  1987.

\bibitem[Xie et~al.(2023)Xie, Gherardi, Pan, and Huang]{xie2023hollownerf}
Xiufeng Xie, Riccardo Gherardi, Zhihong Pan, and Stephen Huang.
\newblock Hollownerf: Pruning hashgrid-based nerfs with trainable collision
  mitigation.
\newblock In \emph{Proceedings of the IEEE/CVF International Conference on
  Computer Vision}, pages 3480--3490, 2023.

\bibitem[Yao et~al.(2020)Yao, Luo, Li, Zhang, Ren, Zhou, Fang, and
  Quan]{yao2020blendedmvs}
Yao Yao, Zixin Luo, Shiwei Li, Jingyang Zhang, Yufan Ren, Lei Zhou, Tian Fang,
  and Long Quan.
\newblock Blendedmvs: A large-scale dataset for generalized multi-view stereo
  networks.
\newblock In \emph{Proceedings of the IEEE/CVF conference on computer vision
  and pattern recognition}, pages 1790--1799, 2020.

\bibitem[Yuan et~al.(2022)Yuan, Sun, Lai, Ma, Jia, and Gao]{yuan2022nerf}
Yu-Jie Yuan, Yang-Tian Sun, Yu-Kun Lai, Yuewen Ma, Rongfei Jia, and Lin Gao.
\newblock Nerf-editing: geometry editing of neural radiance fields.
\newblock In \emph{Proceedings of the IEEE/CVF Conference on Computer Vision
  and Pattern Recognition}, pages 18353--18364, 2022.

\bibitem[Zhang et~al.(2019)Zhang, Jia, Lei, Wang, Ma, and Gao]{zhang2019recent}
Jiaqi Zhang, Chuanmin Jia, Meng Lei, Shanshe Wang, Siwei Ma, and Wen Gao.
\newblock Recent development of avs video coding standard: Avs3.
\newblock In \emph{2019 picture coding symposium (PCS)}, pages 1--5. IEEE,
  2019.

\bibitem[Zhao et~al.(2023)Zhao, Chen, Leng, and Cheng]{zhao2023tinynerf}
Tianli Zhao, Jiayuan Chen, Cong Leng, and Jian Cheng.
\newblock Tinynerf: Towards 100 x compression of voxel radiance fields.
\newblock In \emph{Proceedings of the AAAI Conference on Artificial
  Intelligence}, pages 3588--3596, 2023.

\bibitem[Ziv and Lempel(1977)]{ziv1977universal}
Jacob Ziv and Abraham Lempel.
\newblock A universal algorithm for sequential data compression.
\newblock \emph{IEEE Transactions on information theory}, 23\penalty0
  (3):\penalty0 337--343, 1977.

\end{thebibliography}
}


\end{document}